# Unsupervised Anomalous Data Space Specification

Ian J. Davis
David R. Cheriton School of Computer Science
University of Waterloo, Ontario, Canada N2L 3G1
ijdavis@uwaterloo.ca
(Unpublished paper submitted to https://arxiv.org under Computer Science/Machine Learning)

*Abstract*—Computer algorithms are written with the intent that when run they perform a useful function. Typically any information obtained is unknown until the algorithm is run. However, if the behavior of an algorithm can be fully described by precomputing just once how this algorithm will respond when executed on any input, this precomputed result provides a complete specification for all solutions in the problem domain. We apply this idea to a previous anomaly detection algorithm, and in doing so transform it from one that merely detects individual anomalies when asked to discover potentially anomalous values, into an algorithm also capable of generating a complete specification for those values it would deem to be anomalous. This specification is derived by examining no more than a small training data, can be obtained in very small constant time, and is inherently far more useful than results obtained by repeated execution of this tool. For example, armed with such a specification one can ask how close an anomaly is to being deemed normal, and can validate this answer not by exhaustively testing the algorithm but by examining if the specification so generated is indeed correct. This powerful idea can be applied to any algorithm whose runtime behavior can be recovered from its construction and so has wide applicability.

ACKNOWLEDGMENT

This research was supported by grants from CA Technologies and the Natural Sciences and Engineering Research Council of Canada. Steve Versteeg of CA Labs and Rao Kotagiri's research team at the University of Melbourne, Australia, directed our attention to the significance of previously published research on random forest algorithms. Thanks are also expressed to the ICAC 2015, ASE 2015, ICDE 2016, and PKKD 2016 referees who motivated numerous minor improvements to this paper.

1. INTRODUCTION

The ability to identify anomalous values in arbitrary data without ground truth or external supervision[1] is of considerable importance in noise reduction; in filtering out anomalies as a precursor to data analysis and prediction (Rousseeuw 1984); in responding in real time to anomalous signals that should be addressed or ignored (Thompson 2010); in auditing individuals and corporations; and in discovering concerns that might otherwise go unnoticed. For example, it can be used to discover potentially dangerous medical conditions, fraud, or threats that may cause failure in engineered systems. It can also be used to classify the significance of metrics produced from other sources that themselves classify data.

There are many potential definitions for what constitutes anomalous data (Chandola 2009). When examining an arbitrary sequence of values, the characteristics of the neighborhood in which these values occur may have a bearing on whether they are collectively deemed anomalous (Malik 2014). Conversely, for widely separated time periods, the underlying interpretation of data may change, resulting in corresponding changes as to what is to be deemed anomalous (Gray 2007; Zhou 2009). And depending on the application, the degree to which values must depart from normal, before being deemed anomalous, will vary.

However, if the data of interest (or windowed subsets of this data (Ding 2013; Tan 2009)) can be considered homogeneous, irrespective of the ordering of that data, a simple autonomous presumption about potentially anomalous values, providing that they are not being introduced maliciously, is that they can be expected to be observed infrequently and to be significantly

---

[1] Without semantic data knowledge or definition of anomaly

different from values that are not anomalous. It is this we wish to quantify.

Section 2 presents related work; section 3 outlines the benefits of our proposed algorithms; section 4 explains the basic algorithm; section 5 presents synthetic 1-dimensional results; section 6 generalizes to *n*-dimensions; section 7 presents synthetic 2-dimensional results; section 8 presents real world results; section 9 suggests real world applications; section 10 gives threats to validity; and section 11 provides conclusions.

## 2. PRIOR RELATED WORK

It has been proposed (Liu 2008, 2011, 2012; Ting 2009, 2010) that the unbalanced way in which ordered binary search trees are built can be statistically exploited by using a Random Forest (Breiman 2001) to identify values that are distant from other data and which occur infrequently.

To explain this prior research, if the clustering (Burbeck 2006; 2007) of values in a dataspace is unbalanced, any repeated partitioning of data in that space on some random position (uniformly chosen) strictly within the partitions values will produce unbalanced partitions. Some partitions will contain many values in them, while others will contain few. If partitioning continues until each final partition contains only duplicated values, partitions containing many data points will be partitioned many more times, than those that contain few data points.

This partitioning of a data space can be modelled by an ordered full binary tree. Each internal node represents one such partitioning, and each node's string or numeric key the discriminator on which data under it is partitioned. The depth to any leaf value within this model indicates the number of times a value is partitioned, and thus how "crowded" the partitions are that this value is placed in.

This observation provides a mechanism for isolating the type of anomaly described above from normal data. Simply sample an appropriate number of values without replacement from the provided data, and construct a statistically significant number of trees, using random valued search keys having the property that they split distinct sampled values into two non-empty sub-trees. Such a set of trees is called an isolation forest (*iforest*).

The trained *iforest* can be used in an unsupervised manner to detect whether a value *v* is likely to be anomalous. Compute the number of nodes visited (i.e. the average or cumulative search path length) when *v* is searched for in every tree in the forest. Statistically long search path lengths associate *v* with partitions that contain normal (i.e. densely clustered) data, while short ones associate *v* with partitions more likely to contain rare (infrequently observed and distant values) and thus anomalous data. Thus *v* can be identified as normal or anomalous according to the values it associates with. This effect will be pronounced if anomalous values are rare but present within the sampled data and thus the forest of trees.

Absent such anomalous data the distribution of observed depths is expected to approach the normal distribution as the number of sampled data values becomes large. Consider an ordered binary search tree constructed from a insertion of a random permutation of the values *1,…,n*. The depth $L_n$ of the last value inserted has $mean(L_n) = \Sigma_{i=2,n}\ 2/i$ and variance $var(L_n) = \Sigma_{i=2,n}\ (2-4/i)/i$. As n→∞ this distribution of depths approaches the normal distribution (Devroye 1988). Using Monte Carlo simulation this is demonstrated for 10,000 values drawn from the uniform and exponential distribution (Fig. 1).

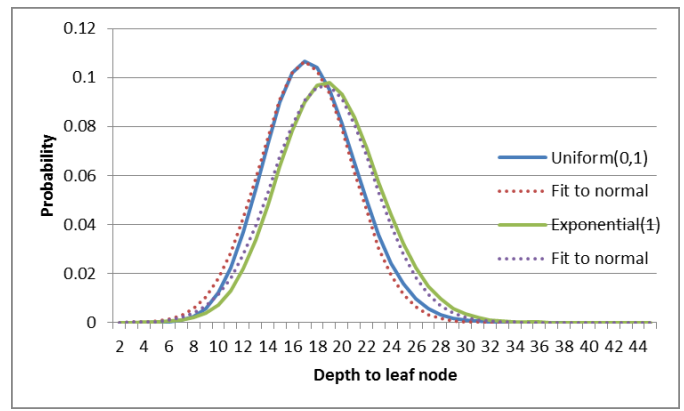

**Figure 1. Probability distribution of search depths**

*C*++ source code for all results presented below is available as open-source code [2], as are the detailed outputs used to generate the results presented here. Earlier *R* source code is also available (Liu 2009).

---
[2] http://cs.uwaterloo.ca/~ijdavis/iforest

The *iforest* algorithm has previously been compared to ORCA, LOF, and Random Forest (Liu 2008, 2012), to SCiForest (Liu 2001) and to SVM (Liu 2012). We have found no algorithms in the literature which generate anomaly specifications, against which we might compare our computed anomaly specifications.

## 3. THIS PAPERS CONTRIBUTION

In order to know which values should be deemed anomalous within an arbitrary data set we must present every value in this data set to the *iforest*. But this *iforest* behaves very much like a black box, and as an external user we are left no wiser as to the logic that is internally employed to discover those data values which are to be deemed anomalous.

Instead of exploiting hidden models to identify anomalies, as previously proposed, we present new algorithms that identify the **precise** rules earlier algorithms employed in isolating anomalous data from more normal data. These rules can be expressed as a sequence of ranges of values that earlier algorithms deem to be anomalous, or more generally in *n*-dimensions as all hyper-rectangles of this *n*-dimensional space within which data points are deemed to be anomalous.

This paper proposes ideas that offer many benefits:

1. Automated *specification* of anomalous data spaces, in addition to *detection* of individual anomalous data values, as an aid to data mining, knowledge discovery, and tool validation.
2. Very fast computation of the areas of a 1-dimensional data space within which points will be reported as anomalous.
3. Construction of this anomalous data space specification uses only a small training set.
4. Generation of exact rules that for both present and absent data distinguish normal values from anomalous ones.
5. The ability to determine how close a value deemed to be anomalous is to being deemed normal.
6. More efficient/immediate validation strategies. Instead of having to first find all anomalies in potentially massive numbers of data points, one can instead simply explore if all easily discovered anomalous ranges are indeed to be deemed anomalous, while all other ranges are not.
7. Tailoring of discovered rules so that engineers can more accurately give instruction as to what really constitutes anomalous data of concern to them and what does not.
8. Substantially improved detection speed. Instead of performing computations of path lengths on many trees for each value presented to the algorithm, a typically small decision tree is pre-computed, which once constructed can immediately distinguish anomalous values from normal.
9. Generating specifications for large numbers of dimensions may prove intractable, but when not, remains a one-time cost. We show that pruning and approximation can be of assistance here.
10. The ability to scale to process significantly larger volumes of data in real time applications.
11. Potentially smaller space requirements for discovered rules than required during precomputation of these rules.

## 4. UNIVARIATE DATA

A set of straight forward enhancements to the *iforest* algorithm permit it to discover precisely those ranges of space within which univariate data will be identified as anomalous.

The role of any constructed tree within the *iforest* is simply to return the search path length associated with a given data value. This binary tree can be transformed into an ordered linked list that achieves the same end. Each node in the linked list merely needs to record the range of data values that (when searched) arrive at each leaf node within the tree and the depth of that leaf node. By traversing the tree backwards, and adding nodes to the head of the linked list, the linked list will have the same final ordering as the binary tree (Fig. 2).

Once all *n* binary search trees have been converted into distinct order linked lists $L_i$ each partitioning the total linear search space into ranges $R_{i,j} \in L_i$ satisfying $R.start_{i,0} = -\infty$, $R.end_{i,j-1} = R.start_{i,j}$ and $R.end_{i,last} = \infty$ the cumulative search path depth of every data range is computed. This linear algorithm computes a new search space $L_{new}$ whose ranges are formed by intersecting all ranges in the original tree search spaces.

```
Crange(Crange **ppHead, double f, double t, int d)
{
  this->from   = f;
  this->to     = t;
  this->depth  = d;
  this->pNext  = *ppHead;
  *ppHead      = this;
}
void
Cnode::ranges(Crange **ppHead, double f, double t)
{
  if (!this->pRight) {
    new Crange(ppHead, f, t, this->depth);
  } else {
    this->pRight->ranges(ppHead, this->split, t);
    this->pLeft->ranges(ppHead,  f, this->split);
} }

Cnode  *pRoot;                  // Root of the tree
Crange *pRangeList = 0;         // Head of the list
…
pRoot->ranges(&pRangeList, -DBL_MAX, DBL_MAX);
```

**Figure 2. Converting an isolation tree to a linked list**

Beginning at $x_0 = -\infty$ we compute the ordered sequence of distinct values $x_k \in \{R.end_{i,j} : \forall L_i \wedge R_{i,j} \in L_i\}$ ending at $x_{last} = \infty$. $L_{new} = \{R_{new,k} : R.start_{new,k} = x_k \wedge R.end_{new,k} = x_{k+1}\}$. Summing the path depths of all old $R_{i,j}$ that intersect each $R_{new,k}$ gives the cumulative path depth for this $R_{new,k}$.

Unfortunately, except in simulations, the number of anomalies present in data is unknown. However, it is known that anomalous data will have average path lengths considerably shorter than that associated with typical data (Mann 1947; Sawilowsky 2002).

While these path lengths will, in practice, vary we can still attempt to identify where this divide occurs by ordering all computed path lengths in $L_{new}$ and searching for the largest separation between these values. If the examined distribution appears normal, with no such large gaps between consecutive depths we can conclude that there are no observed anomalies.

The path lengths of data points may alternatively be examined, when not wishing to perform anomaly detection on real time streamed data. This is likely to be less efficient, and may deliver different results, since some ranges of values in $L_{new}$ may not be present in the examined data.

A greedy algorithm beginning at both ends of the sorted (Davis 1992) data advances these two positions toward each other according to which ever has the smallest difference between its depth and the one to be advanced to. Both positions advance on ties. Where they meet is a reasonable guess as to the depth of the anomaly with the longest search path, and by examination (or sorting) thus the number of anomalies in the observed data.

We now have a cumulative boundary depth that is presumed to divides anomalies from normal values. Each computed range $R_{new,j}$ having a cumulative path depth less than or equal to this, is added to the list of anomalous ranges (Fig. 3).

```
Canomaly **ppTail, *pHeadAnomaly, *pAnomaly;
Crange    *pRange;
int       tree, depth;
double    f /* from */, t /* least to */;

pHeadAnomaly = pAnomaly = 0;
ppTail       = &pHeadAnomaly;
for (f = t = -DBL_MAX; t != DBL_MAX; f = t) {
  t = DBL_MAX;
  for (tree = 0; tree < FOREST_SIZE; ++tree) {
    pRange = lists[tree]; /* Head of list */
    if (pRange && pRange->to < t) {
      t = pRange->to; /* Least remaining to */
  } }
  depth = 0;
  for (tree = 0; tree < FOREST_SIZE; ++tree) {
    pRange = lists[tree];
    if (pRange) {
      depth += pRange->depth;
      if (pRange->to == t) {
        lists[tree] = pRange->next;
        delete pRange;
  } } }
  if (depth <= LARGEST_ANOMALY_DEPTH) {
    if (pAnomaly && pAnomaly->to == f) {
      pAnomaly->to = t; /* Just extend */
    } else {            /* Add to list */
      pAnomaly = new Canomaly(f, t);
      *ppTail  = pAnomaly;
      ppTail   = &pAnomaly->pNext;
} } }
```

**Figure 3. Computing anomalous ranges**

**Lemma 1:** If $C$ is the intersection of ranges in $A$ and $B$, which cover the same space, without overlap, then $|C| \le |A| + |B|$. Consider a sequence of contiguous ranges $R_{0...}R_{j-1}$ in $A$. $R_0$ will intersect with $n$ ranges in $B$, producing $n$ new ranges. All but the last range will be subsumed and thus discarded, as will $R_0$. So $n$ ranges are created and $n$ are discarded. Assume this is also true for $R_j$. Then $R_{j+1}$ intersects $m$ regions, all but one being discarded as is $R_{j+1}$. The proof is by induction.

The proposed algorithm is thus efficient and terminates. Having produced an ordered linked list all of the anomalous ranges within the data, this can now be converted back into a balanced search tree to permit rapid searching (Fig. 4).

```
Canomaly *Canomaly::to_tree(void)
{
  int cnt, half;
  Canomaly *p, *next;

  cnt = 0;
  for (p = this; p; p = p->next) ++cnt;
  half = cnt >> 1;

  for (p = this; half; p = next) {
    next = p->next;
    if (!--half) {
       p->next = 0;
  } }
  if (this != p) {
    p->left = this->to_tree();
  }
  if (p->next) {
    p->right = p->next->to_tree();
  }
  p->next  = 0;
  return p;
}

tree = head_anomaliesP->to_tree();
```
**Figure 4. Creating an ordered search tree of anomalous ranges**

## 5. ONE DIMENSIONAL RESULTS

For most applications the *iforest* algorithm detects anomalies efficiently by being trained on a small sample of the data values within the overall data space. It has earlier been suggested that 100 binary search trees each constructed from 256 random data values (selected without replacement but potentially duplicated) is a reasonable training set from which to obtain statistically significant results (Liu 2012).

We implemented precisely this proposed algorithm and applied it to 5,000 generated data points, which have anomalies occurring ~0.5%, ~1.0% and ~2% of the time. These anomalies are assigned ranges of values that are adjacent to but do not overlap normal values. Such anomalies are indeed identified with near perfect precision and recall simply by selecting the same number of data values as anomalies introduced, in order of increasing observed cumulative search path length.

**Experiment 1:** 200 simulations are performed of the *iforest* algorithm on 5,000 data values randomly constructed so that ~99% of the values are in the normal ranges (-1.0, -0.5) or (0.5, 1.0) with equal likelihood. The remaining ~1% of the values are in the range ±1.5 but outside the normal range (Table 1). It is hoped that the *iforest* algorithm will deem this ~1% of the data anomalous, and so produce rules that closely approximate the rules used when introducing these anomalies.

**Table 1. Anomalous ranges**

| -1.5,-1.0 | -0.5,0.5 | 1.0, 1.5 |
|---|---|---|

Results are presented in Table 2. The rows show consecutive results obtained from repeating this experiment. The first column in Table 2 indicates the number of anomalies that the modified *iforest* algorithm identifies as present, and in parenthesis the number actually introduced by the simulation.

**Table 2. Sample rules produced by Experiment 1**

| Guess | Anomalous data ranges | | | | |
|---|---|---|---|---|---|
| 54 (54) | -∞,-1.00 | -0.41,0.38 | 1.04,∞ | | |
| 55 (40) | -∞,-0.99 | -0.45,-0.45 | -0.45,0.41 | 1.01,∞ | |
| 43 (43) | -∞,-1.07 | -0.35,0.33 | 1.11,∞ | | |
| 59 (59) | -∞,-1.03 | -0.37,0.39 | 1.03,∞ | | |
| 40 (40) | -∞,-1.09 | -0.39,0.36 | 1.05,∞ | | |
| 55 (55) | -∞,-1.09 | -0.36,-0.36 | -0.36,0.37 | 1.11,∞ | |
| 43 (43) | -∞,-1.09 | -0.40,-0.40 | -0.40,-0.40 | -0.4,0.37 | 1.05, ∞ |
| 54 (54) | -∞,-1.07 | -0.37,0.40 | 1.09,∞ | | |
| 53 (53) | -∞,-1.09 | -0.35,0.39 | 1.09,∞ | | |
| 52 (52) | -∞,-1.10 | -0.30,0.39 | 1.08,∞ | | |
| 53 (53) | -∞,-1.06 | -0.40,0.37 | 1.08,∞ | | |
| 47 (49) | -∞,-1.04 | -0.34,0.38 | 1.06,∞ | | |
| 59 (60) | -∞,-1.04 | -0.37,-0.37 | -0.36, 0.37 | 1.03,∞ | |
| 45 (47) | -∞,-1.09 | -0.33,-0.33 | -0.33,-0.32 | -0.32,0.25 | 1.05,∞ |
| 13,(39) | -∞,-1.24 | | | | |

As can be seen, the decision rules for identifying anomalies in each experiment is concise, and for the most part closely approximates the rule used to introduce anomalies.

The union of the ranges identified as anomalous in the 15 experiments shown in Table 2 identifies anomalies with a precision of 92.5%. This demonstrates how quickly and accurately the *iforest* algorithm can discover hidden rules employed in seeding anomalous data, with only minimal computation (Table 3).

**Table 3. Union of anomalous rules shown in Table 2**

| -∞,-1.00 | -0.45,0.41 | 1.01, ∞ |
|---|---|---|

**Experiment 2:** 200,000 values are generated with normal values uniformly distributed in the range ± (0.5, 1.0) and anomalous values in the range ± (2.0, 100). Independent experiments are conducted twice on each of the following anomaly distributions {0.005%, 0.01%, 0.02%, 0.05%, 0.1%, 0.2%, 0.5%, 1%, 2%, 5% and 10%}. The number of anomalies introduced into each experiment is then predicted and compared to the number of anomalies introduced (Fig. 5).

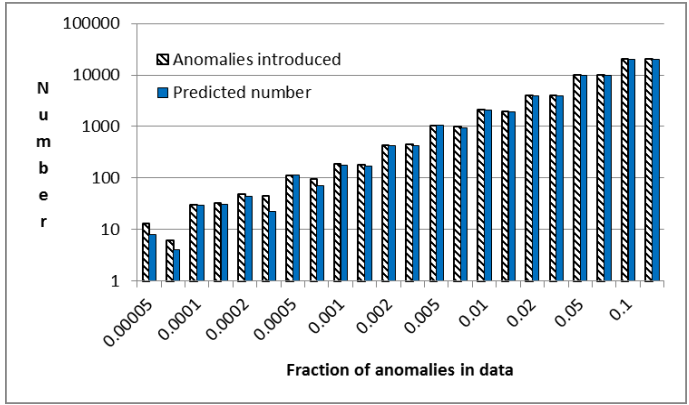

**Figure 5. Predicting the number of anomalies in data**

## 6. MULTIVARIATE DATA

Normal behaviour is often the consequence of complex relationships between multiple variables. For example, we may wish to study how anomalies in data values vary over time. To detect anomalies in multivariate data the strategy for discovering anomalous ranges in one dimension is generalised to an algorithm that can also be applied to multivariate data in $n$ dimensions. To explore the feasibility of such a generalisation, the basic algorithm is described that discovers anomalous regions in two dimensions.

Each binary tree is replaced by a KD-tree (Bentley 1975), in which at each node the variable that directs the search path alternates (while possible) between the two dimensions. Using the KD-tree the search space that documents the leaf at which every coordinate arrives for each tree in the forest is computed, as is the depth of this leaf. For any given tree the rectangular spaces associated with each leaf collectively cover without overlap the two-dimensional data space.

**Lemma 2**: The intersection $C$ of two search spaces $A$ and $B$ each containing only disjoint regions, also contains only disjoint regions. For if a single region in $A$ intersects regions in $B$, the result remains disjoint. Suppose this is also true for $k$ regions in $A$. Then the result remains disjoint. Intersecting one further region with a disjoint set produces a disjoint set. So this is also true for $k+1$ regions. The proof is by induction.

**Lemma 3**: The intersection of two search spaces $A$ and $B$ that cover the same space $S$, results in a search space $C$ that also covers (without overlap) exactly $S$. For if any point in $S$ was absent in $C$, it would not be subsumed by a region in both $A$ and $B$. But both $A$ and $B$ cover S. Proof by contradiction.

Each KD-tree in the forest generates search space rules, which define how that KD-tree is searched. We wish to compute all possible intersections of this large set of search spaces, while summing the depths of all regions that intersect, producing a final search space. This final search space can then be used to describe the cumulative search path depth across all KD-tree's in the forest for any Cartesian coordinate.

In one dimension, if there are $k$ search keys in the *iforest* the data space will be partitioned into $k+1$ regions. These regions can be examined in some small constant time or small linear time w.r.t. *iforest* size. In two dimensions the data space is instead partitioned into rectangles on the two orthogonal search dimensions. In $n$-dimensions the data space is partitioned on every orthogonal search dimension into hyper-rectangles. The number of such hyper-rectangles is thus the product of the number of partitions in each dimension. So in general while $n$-dimensional detection remains linear w.r.t. the size of the data being examined, specification requires examining $O(c^n)$ hyper-rectangles where $c \approx k / n$. For high dimensionality the examination of all these hyper-rectangles rapidly becomes intractable.

In 2-dimensions specification remains feasible, since in our experiments $c \approx 12,800$ and in 3-dimensions perhaps still justifiable as a one-time cost.

Pruning the *iforest* can significantly speed computation of anomaly specification when performance is of concern. Most obviously, if an explicit path depth is associated with each node, those nodes for which all paths through them to leaf nodes have the same depth $d$, can be associated with this depth $d$ and all subordinate nodes then ignored.

If an upper bound on the anomaly cut off depth can be obtained, or an anomaly cut off depth chosen without considering the path depth associated with every hyper-rectangle, those nodes under which all data points are normal can be efficiently identified. One such obvious cut off is the average summed path length, since it would be unreasonable to expect cumulative anomalous path depths to exceed this average cumulative depth, but tighter bounds can probably be determined through appeal to the maximum number/percentage of anomalies anticipated in the data, and the fact that path lengths are expected to be distributed approximately normally.

For increasing path depth, compute the hyper-rectangle describing the data values that can arrive at each internal node in the *iforest*. If this depth plus the sum of all other minimal depths through all other trees employing any value contained within this same hyper-rectangle exceeds the chosen anomaly cut off depth, this hyper-rectangle cannot contain any anomalous data points. Consequently again all nodes beneath it can be ignored, and this node treated instead as a leaf node.

To improve performance further, while reducing memory usage, it can be recognised that anomaly specification at best approximates ground truth. Consequently boundaries in any one dimension can be centered between the desired precision of the data or result, (minimizing intersections) and/or very narrow partitions can be merged with the next partition in that dimension. In particular, if results are being presented visually, as they are here, hyper-rectangles having any dimension finer than the chosen image resolution can be merged with the next.

Pixelation is used to construct the anomaly specification. For 2-dimensional data we first build a sorted *x* and *y* vector each referencing all *r* rectangular search spaces in all trees within the forest. If we have 100 forests, each containing 256 leaf nodes, each vector has at most 25,700 entries. Entries in the *x* vector are sorted by the right edge of the search space referenced, and the *y* vector by the top edge. There is also a matrix *pixel*. Each *pixel[i][j]* contains the computed average search path depth for the search space having edges *x[i-1].right*, *x[i].right*, *y[j-1].top*, and *y[j].top*. As special cases *x[-1].right* evaluates to *x[0].left*, and *y[-1].top* to *y[0].bottom*, and pixels having no area are eliminated.

A KD-tree efficiently computes cumulative search path depths. The KD-tree indexes the *r* search spaces by discriminating repeatedly on their four corner coordinates. Each leaf node references the search spaces thus indexed. We present each pixel to this KD tree, which locates all search spaces that contain this pixel. This involves searching all paths within the KD-tree that can index search spaces containing this pixel, while ignoring those that cannot. The pixel is then assigned the summed search path depth of these located search spaces.

Having computed the search path depth of every pixel, pixels can be identified as normal or anomalous according to their depth. To concisely represent anomalous regions, anomalous pixels are converted into larger rectangular search spaces. This can be achieved by using any suitable algorithm. We use a greedy algorithm. We find the anomalous pixel having smallest left-bottom coordinate. We enlarge this to the largest square containing only anomalous pixels with this left-bottom pixel, and repeat until all of the anomalous pixels have been thus combined. We then combine rectangles.

Having converted overlapping anomalous spaces into disjoint sub-spaces, anomalous spaces can then be efficiently located using KD-trees, quad-trees (Finkel 1974), etc. This trained model can then be efficiently accessed to identify anomalous values, without having to compute path lengths for every data value.

## 7. TWO DIMENSIONAL EXPERIMENTS

The same basic parameterization is used in two dimensions as in one. We again examine 5,000 data points. Each forest contains 100 trees, each KD-tree ≤ 256 leaves, and ~1% of the data randomly generated is anomalous. This probability, but not the number of anomalies introduced is known to our algorithm. Pixels with width or height < 0.005 are eliminated to improve performance.

**Experiment 3:** (x,y) coordinates are randomly uniformly generated in the range ±1. ~1% of coordinates are assigned an anomalous value in the range ±2 but outside the range ±1. Anomalous values are identified with a precision of 0.94, recall 0.85 and the f-measure 0.90. The computed anomalous search space (the central focus of this paper) is highlighted in pale yellow (Fig. 6).

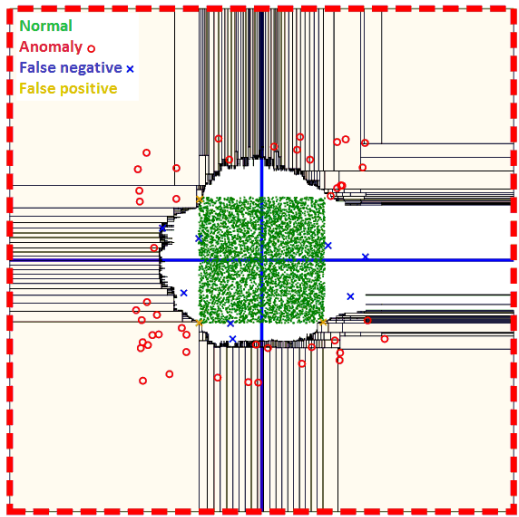

**Figure 6. Anomalies outside the box**

In the absence of any external sampled data points, partitions containing sampled edge and corner points will extend to ±∞. Consequently huge search spaces can be specified arbitrarily as normal or anomalous, merely because of their proximity to such dominant edges and corners.

Few sampled points exist in the region of a corner, and so these points are likely to be erroneously specified as anomalies. This in turn may cause rejection of an equivalent number of anomalies because they consequently have a larger depth ranking. Of course, knowing nothing about the nature of anomalous data, it may be reasonable to presume that boundary values we might deem normal, are instead abnormal.

The number of sampled points near a long edge is typically larger, causing their search paths to be longer. Thus they are less likely to be deemed anomalous.

**Experiment 4:** (x,y) coordinates are rotated 45 degrees. Anomalous values are now identified with a precision of 1.0, recall 0.95 and f-measure 0.97 (Fig. 7). This demonstrates the degrees to which data edges that align with partitions, tend to impede *iforest* anomaly detection, and suggests that better results may be obtained by describing anomalous spaces under a variety of data rotations.

**Experiment 5:** Results may also be improved by independently computing anomalous regions repeatedly, and then forming the union of these anomalous regions. The result of computing all anomalous data regions 5 times and then visually overlaying all such regions is shown (Fig. 8).

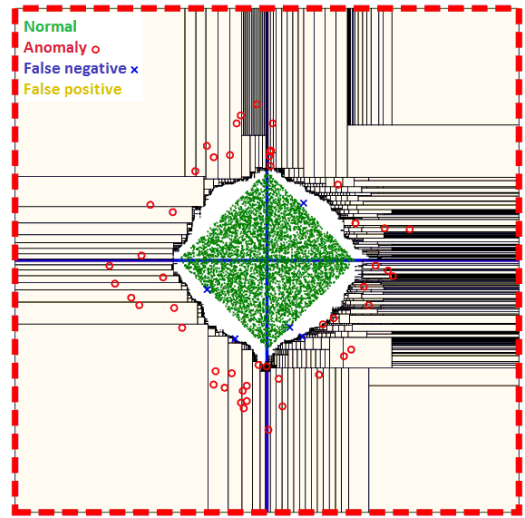

**Figure 7. Anomalies outside the diamond**

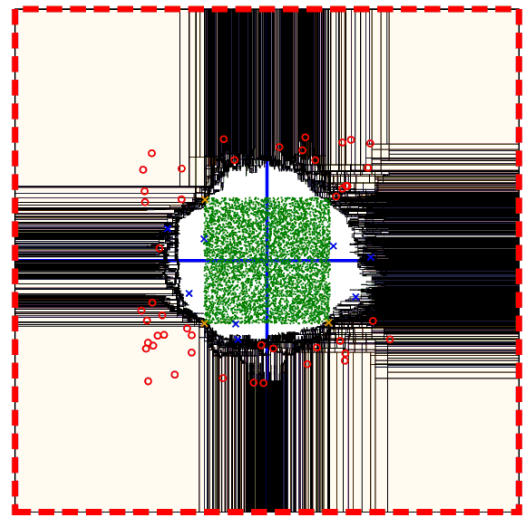

**Figure 8. Anomalous regions from 5 experiments**

**Experiment 6:** (x, y) coordinates are randomly generated in the range 0 to ±2 with all normal coordinates having the same sign. ~1% of coordinates are randomly assigned an anomalous difference in their sign.

Over 100 random experiments, both precision and recall was 46 percent. This result seems poor compare to the earlier results we observed. To help explain this behaviour, the partitioning generated by one tree in the forest is shown (Fig. 9).

Note that partitioning is a consequence of sampled data, and not the collective data shown. As can be seen, there is a strong tendency for partitions containing boundary data points to extend into anomalous regions, and as noted earlier this effect impedes anomaly detection.

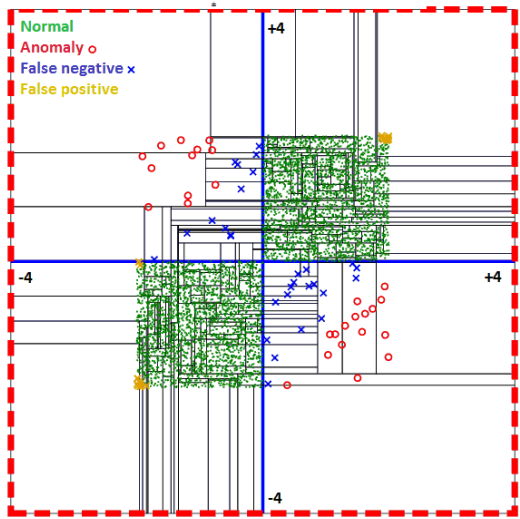

Figure 9. Anomalous anti-correlated sign

**Experiment 7:** To improve separation, experiment 6 was modified so that no data coordinate had a value in the range ±0.0-0.5. This significantly increases the separation between normal and anomalous values. The resulting precision and recall following this change was much better at 88% (Fig. 10).

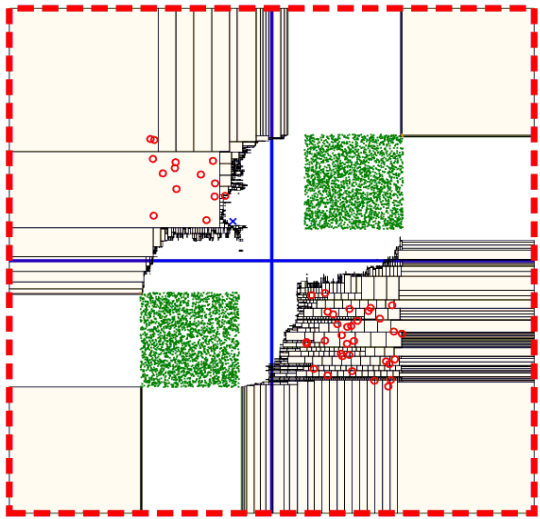

Figure 10. The anomalous search space

**Experiment 8:** Experiment 7 is repeated, but with all values $v$ being replaced by $1/v$, thus distributing values according to the inverse uniform distribution. This creates significantly skewed data values and as consequence results in precision 0.12, recall 1.0 and f-measure 0.21 (Fig. 11).

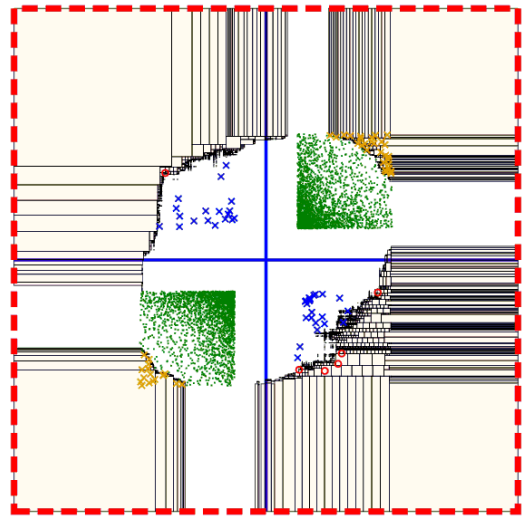

Figure 11. Using inverse uniform distribution

**Experiment 9:** In this experiment, everything between a unit 1 and 2 circle is normal, and everything else anomalous (Fig. 12). Because the entirely unsupervised algorithm naturally tended to partition the inside of the circle, but not all regions of the outer edges of normal data, precision was 0.33 and recall 0.38. Most anomalies within the unit circle (~50%) failed to be detected, demonstrating that the *iforest* algorithm can completely fail to identify some types of anomaly, even if transformed through rotation.

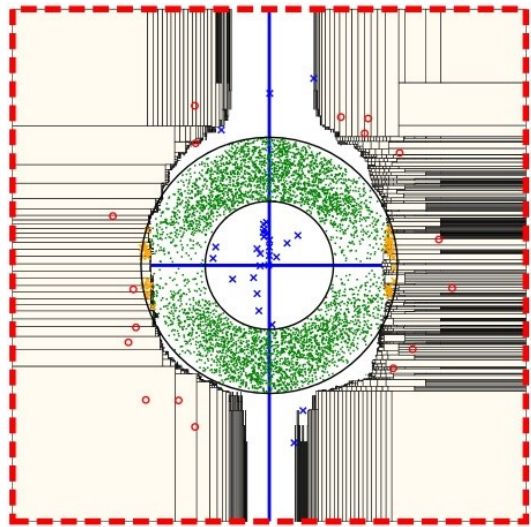

Figure 12. Anomalous values lie outside the ring

8. REAL WORLD EXPERIMENTS

The behaviour of the isolation forest is now examined on CPU load averages obtained from a large cloud environment. The data examined spanned 7 to 12 years; it was collected hourly, and was maintained separately for each cloud client. Parameterisation is largely unchanged. However, given the volume of data examined, the sampled values per tree are increased from *256* to *512*. This results in better differentiated path lengths. The trees per forest remains set at *100*.

Client data from the *5,858* distinct time-series are processed sequentially and independently. Zeroes are discarded and 3,781 time series too short to be sampled are ignored. Collectively, 42,664,075 useful data points are examined. By time-series this number has a minimum (*min*) of *512*, a mean ($\mu$) of 20,541 a maximum (*max*) of *159,250* and standard deviation ($\sigma$) of *29,591*.

**Experiment 10:** All 42,858,960 data points are treated as being derived from a single homogenous set, within which anomalies are specified and detected. 1,662 secs were spent reading this data; 0.02 secs to build the single *iforest*; 0.0005 secs to estimate maximum cumulative anomaly depth; 0.01 secs to compute the anomalous data space specification; and 2.1 secs to detect all anomalies using this specification; or alternatively 132 secs to detect anomalies directly from the *iforest* without using the precomputed specification.

**Experiment 11:** Each time-series is processed separately. 1,821 secs were spent reading data; 24 secs to build all the *iforests*; 0.5 secs to estimate all the maximum cumulative anomaly depths; 11.4 secs to compute for each time-series the anomalous data space specification; and 2.5 secs to detect anomalies using these specifications; or alternatively 132 secs to detect anomalies without using the precomputed specification.

**Experiment 12:** The number of predicted anomalies within each time series is determined. For each time series, the isolation forest algorithm independently estimates the number of anomalies present five times. The outlier farthest from the mean is removed, and the mean of the closest remaining four predictions computed. Maximal change in predicted anomalies is less than ±5% and the average is less than ±1% (Fig. 13).

This experiment took 2,353 secs. 1,156 secs were spent reading data subsequently used; 547 secs data subsequently ignored; 641 secs building *iforests* and computing training depths 5 times; 9.08 secs sorting depths (using a linear radix sort) 5 times; and 0.12 secs to then computing the number of predicted anomalies (now occurring at the start of the sorted list) 5 times.

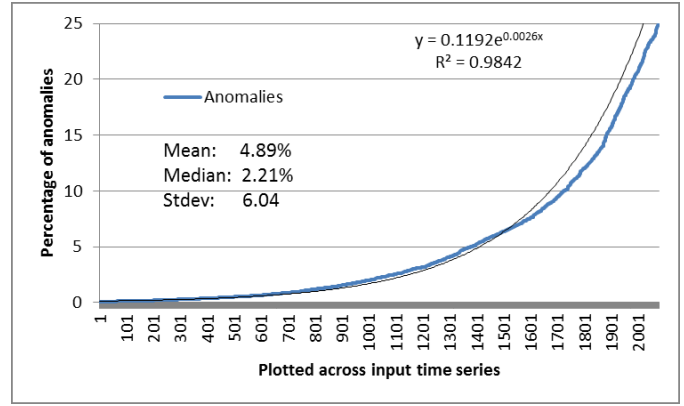

**Figure 13. Estimated anomalies in data by input source**

It is difficult to verify that the isolation forest algorithm correctly identifies anomalous data spaces in large data sources without an in-depth understanding of the data examined and the true nature of anomalies within it. Unsupervised testing is therefore employed, in which the algorithm validates itself, with no external specification about the nature of anomalies.

**Experiment 13:** For each client input time series the isolation forest algorithm is randomly executed twice. The first execution provides a control, and establishes quasi-anomalies within the data. The second execution is evaluated on its ability to identify these same anomalies, from independently sampled data and an independently constructed forest. If a small sample of the data initially identified as anomalous is then similarly specified by a second independent process, something within this data must distinguish it from the vast majority of values, and to that extent at least this data can be labelled anomalous.

Averaged statistics across inputs are presented for maximum estimated anomaly depth (*a*), minimum tree depth ($d_{\geq}$), average depth ($d_{\mu}$), maximum depth ($d_{\leq}$), precision (*p*), recall (*r*) and f-measure (*f*). Since the isolation forest algorithm operates by ranking its input, the Spearman's Rank Correlation coefficient ($\rho$) is computed on the ranking of depth for each data point over the two executions. The average precision and recall is 85% and there is near perfect correlation at *93%*. (Table 4).

**Table 4. Real world results**

|   | $a$ | $d_{\geq}$ | $d_{\mu}$ | $d_{\leq}$ | $p$ | $r$ | $f$ | $\rho$ |
|---|---|---|---|---|---|---|---|---|
| min | 3.94 | 2.16 | 11.89 | 13.00 | 0 | 0.01 | 0.01 | 0.51 |
| $\mu$ | **9.66** | **4.79** | **13.48** | **15.91** | **0.85** | **0.86** | **0.79** | **0.93** |
| max | 18.39 | 13.0 | 18 | 21.83 | 1.00 | 1.00 | 1.00 | 1.00 |
| $\sigma$ | 2.56 | 1.12 | 0.97 | 1.51 | 0.27 | 0.25 | 0.28 | 0.07 |

**Experiment 14**: Running the *iforest* algorithm independently on each time series, without domain knowledge or external supervision, we compute the specification of anomalous CPU load ranges presuming that 0.2% of data is anomalous. The frequency with which a value is deemed anomalous across all these independent specifications is then shown as a percentage (Fig. 14). Values larger the range shown are also anomalous. However only 17.5% of the generated algorithms considered negative values anomalous. While unlikely to be discovered during testing, this flaw once identified in a specification can be easily corrected.

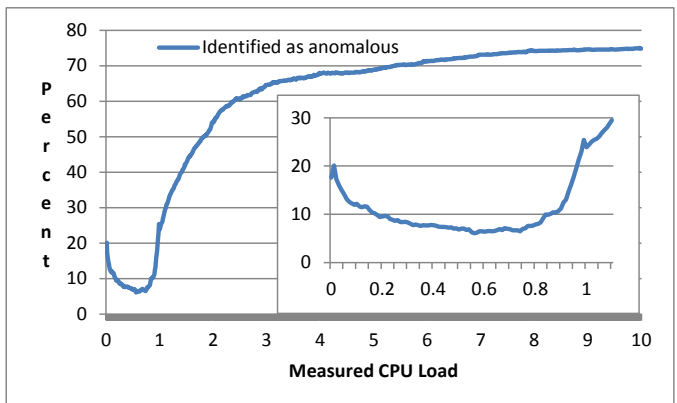

**Figure 14. Identifying anomalous CPU load data ranges**

Once the nature of anomalies in the data has been determined, data ranges can be identified that with high probability are genuinely anomalous. Armed with this knowledge, experiments can be conducted on real world data, either presumed to contain a known percentage of anomalies, or examined to determine the number of anomalies within it. Statistics for path depths in forests derived from data containing such anomalies (such as those shown in Table 4) can then be computed. These statistics can be referenced when seeking to discover anomalies in new data, and by being suggestive used to provide better estimate of the expected number of anomalies in this new data.

## 9. PRACTICAL APPLICATIONS

As noted in the introduction, there are many practical applications for anomaly detection. However, it is unclear how effective the proposed algorithm will be when implemented within real world applications, primarily because it presumes that data is homogeneous, and that the order in which data appears within a time series is irrelevant. We now explore this issue.

### 9.1 Discontinuity detection

It has earlier been proposed that anomalies and discontinuities in large-scale systems be discovered by seeking points in the time-series where the locality is least smooth (Malik 2014). One efficient way of seeking the position of such a discontinuity is to partition each time series into two independent subsequences $S_1=\{v_1...v_{i-1}\}$ and $S_2=\{v_{i+1}...v_n\}$ with $v_i$ chosen so that $LSE(S_1)+LSE(S_2)$ is minimized.

**RQ1**: Are the values $v_i$ where a time-series is least smooth contained within the anomalous data regions identified by the proposed *iforest* algorithm? If so discontinuities in the data may be consequence of (or signaled by) anomalous values, and might be more efficiently located by first examining the region of a time series where such anomalies are detected.

When it is assumed that 1.0% of the values present are anomalous 45% of the $v_i$ are identified as anomalous. When the algorithm autonomously estimates the number of anomalies present, the ratio of anomalies detected rises to 5.57% (distributed as earlier shown in Fig. 13), and the percentage of $v_i$ deemed to be anomalous rises to 56%. Thus the *iforest* algorithm can be used to discover discontinuities in about half the cases, but time series order and local context remains significant in identifying those discontinuities not associated with any discovered anomaly.

Generalising, if $k$ anomalies are detected in a time-series, what percentage of the $k$ most significant outliers fall within this set of anomalies, if the significance of an outlier is determined by its distance (*LSE)* from the linear regression line computed across all the data in a given time-series (Fig. 15).

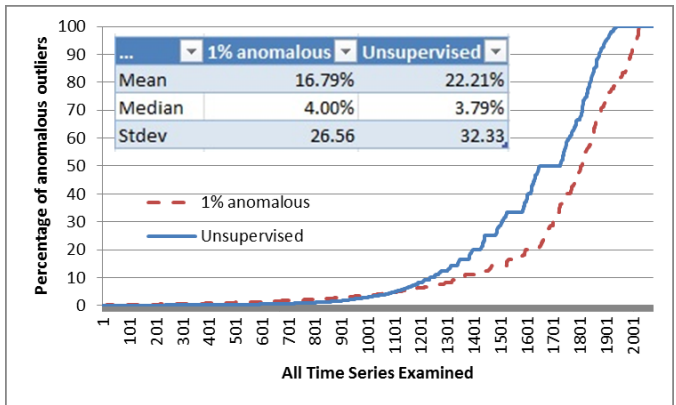

Figure 15. Overlap between anomalies and outliers

The overlap between identified anomalies and outliers by frequency, when the number of anomalies is unsupervised and outliers are chosen to agree with the number of automatically discovered anomalies in each time-series is shown in Fig. 16.

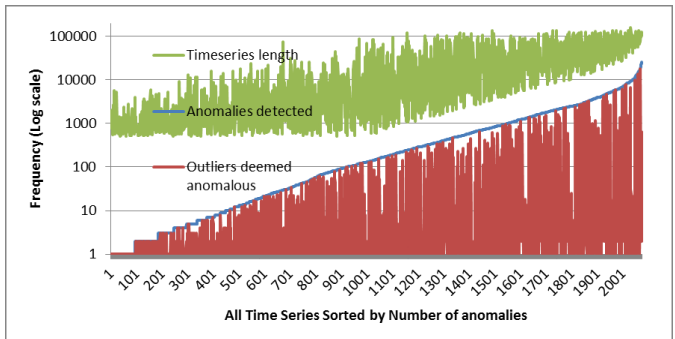

Figure 16. Anomalies and outliers deemed anomalous

## 9.2 Multivariate linear regression

Multivariate linear regression is not robust when presented with outliers, since a single outlier can have an unbounded effect on the least squared estimator (Salibian 2006).

**RQ2**: Are the predictions of multivariate linear regression improved by removing from the input time series those values specified by the extended *iforest* algorithm as being within an anomalous data space? Prediction should be improved if these removed values really are anomalous outliers.

A previously described cloud environment is used to investigate this question (Davis 2013). As described in that paper multivariate regression working with hourly data attempts to predict the next hours transaction processing power (TPP) by employs lags in utilization of 1 and 2 hours, 1 and 2 days, 1 through 4 weeks, and when possible 1 and 2 months.

There are many missing values within this data. Any time series that contain more interleaved missing values than useful data is discarded as useless, as are very short time series. Otherwise, interleaved missing values are assigned the same value as the last present value.

The probability density function (PDF) for observed absolute residues (not associated with missing values) is approximated using a histogram. This is done separately for varying choices as to the number of most anomalous values to remove from the data. Figure 17 shows the resulting shift in the PDF when these anomalous values are treated as missing values, instead of being treated as valid data points.

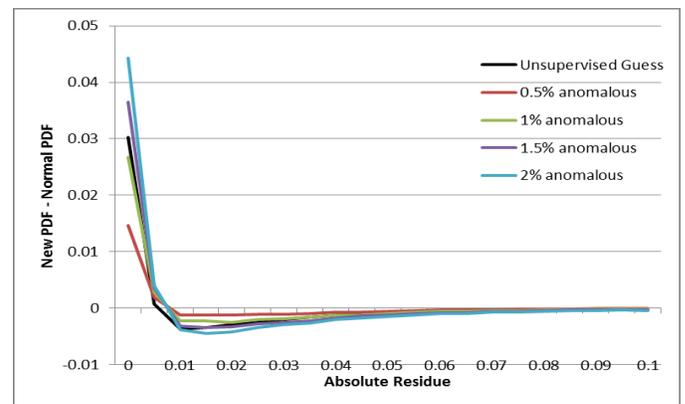

Figure 17. Improvement when removing anomalies

As expected, removal of anomalous values shifts the PDF so that very small residues become more likely, while larger residues become less likely. As more anomalous values are treated as missing, this shift becomes more pronounced. Thus prediction using regression is improved for remaining values.

## 9.3 Anomaly detection in multiple dimensions

**RQ3**: It is important to be able to identify anomalies within images. We demonstrate 3-dimensional specification and detection by finding unusual {R,G,B} colors in an image. Since values are integer, total partitions are minimised without other effect by setting all search keys $k_i =[k_i]- 0.5$.

A 700 x 555 image[3] (Fig. 18) is read by *libnsbmp*[4] and all except those 93,919 color pixels autonomously identified as anomalous (24%) are replaced with white, visually highlighting the anomalous colors (Fig. 19). The specification of anomalous colors is shown in Table 5. As explained in experiment 4, the RGB color cube is rotated 45° in the horizontal and vertical plane, prior to processing, to avoid corners of the RGB cube wrongly being treated as anomalous.

---
[3] www.esa.int/spaceinimages/Images/2015/12/The_Lockman_Hole_in_X-rays
[4] http://www.netsurf-browser.org/projects/libnsbmp

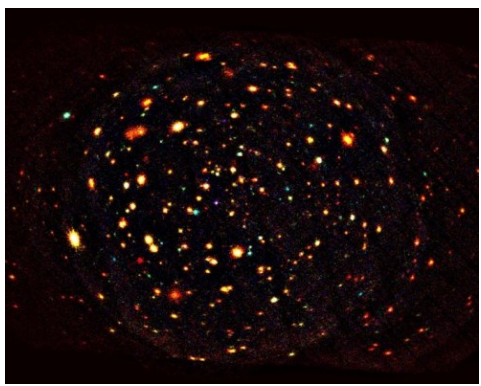

**Figure 18. The Astronomic Lockman hole in X-Rays**

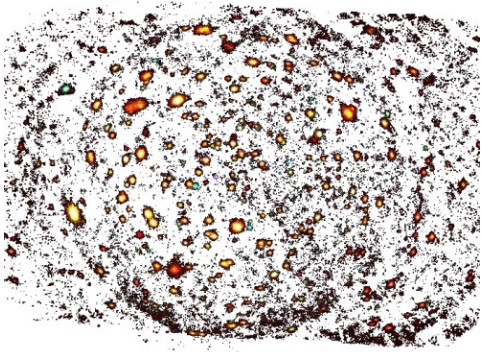

**Figure 19. Anomalous color points within this image**

The *iforest* was built in 0.79 secs; detection without prior specification took 1.11 secs; and computation of the specification took 8.14 secs.

**Table 5. The two identified anomalous color spaces**

| From | To | Spectrum |
|---|---|---|
| {00,85,99} | {FF,93,A3} |  |
| {00,00,1A} | {FF,73,8C} |  |
| Normal complimentary color space |  |  |

### 9.4 Data Analysis

**RQ4**: For a horizontally scalable cloud service what would constitute anomalous pairings of transaction frequency and response time, and how anomalous would these pairings be?

Transaction frequency is captured every 15 minutes for an arbitrary service running thoughout December 2014, and average server response time during each of these 15 minute intervals is correlated. A contour map is presented showing for every possible coordinate, what percentage of the observed data coordinates (if deemed normal shown in black else red) have shorter cumulative iforest path lengths than this coordinate does (Fig. 20).

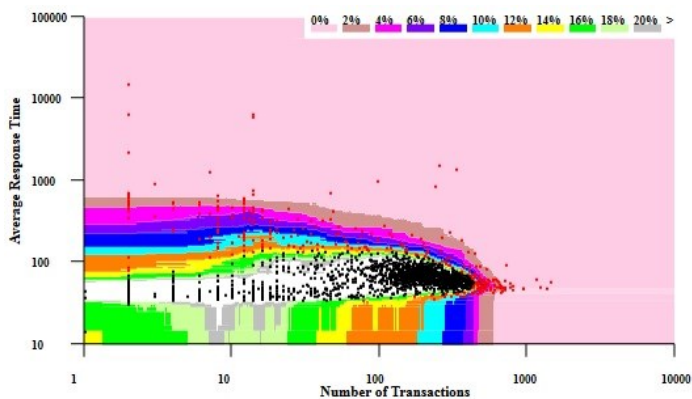

**Figure 20. Response times versus transaction frequency**

An analyst when presented with such a contour map can then easily modify it to indicate the space of anomalies which should trigger intervention, even if they were not earlier familiar with the operational characteristics of the system. In the above example those data points indicating that excessive numbers of transactions do not typically increase the average response time suggests that load balancing is working remarkably well. However, those data points associated with poor response times, particularly under light loads, should generate some sort of automated alarm.

In Fig. 17 chart resolution was 450 pixels by 250 pixels and at that resolution contour maps can be computed in approximately 0.1 secs per time-series. 311 anomalies (15%) were detected in the 2081 data points. This chart was not generated from a specification, but by computing path lengths for a representative (central) value in each pixel.

### 9.5 Comparative Algorithm Analysis

**RQ5:** To what extent do the red anomalies shown in Fig. 20 correspond to *k*-nearest neighbor distance based outliers? (Breunig 2000; Knorr 1998, Ramaswamy 2000)

For each data point $p_i$ the distance $d_k$ to the $k$ nearest neighbours is computed, and data points are then ordered by descending $\max(d_k^2/k : 1 \leq k \leq 200)$, this being inversely proportional to population density within the circle of radius $d_k$ centered at $p_i$. Earlier ordered data points are thus surrounded by some region that is less densely populated than any discovered region associated with a later data point, and are thus ranked as more significant NN-outliers.

When $1 \leq k \leq 200$ varies independently for each data point f-measure is maximal for the 297 most significant NN-outliers, matching 253 known anomalies, with precision 0.852, recall 0.814 and f-measure 0.832. It is a consequence of the definition of precision, recall and f-measure that the number of anomalies will necessarily be in agreement with the number of outliers where these three values intersect. That is at 0.823. Pearson's $r$ for the two rankings schemes is 0.742 (Fig. 21).

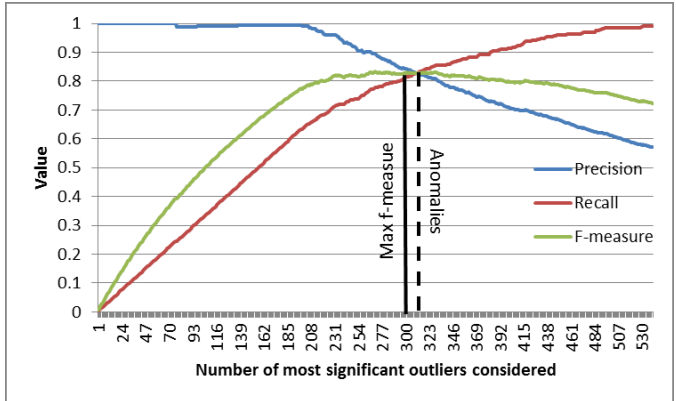

**Figure 21. Comparison to *k*-nearest neighbour detection**

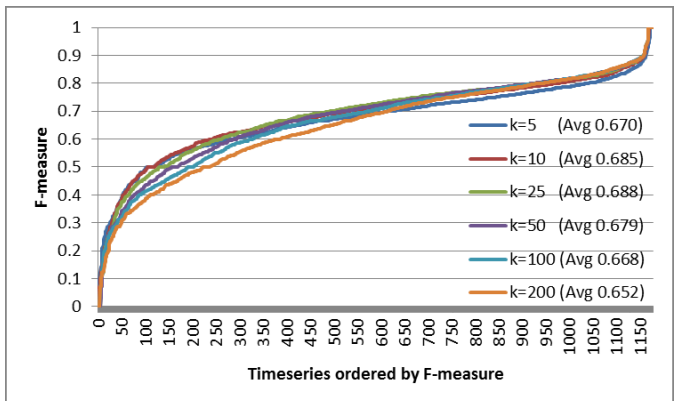

**Figure 22. *Iforest* agreement to variable *k*-nearest neighbor**

The average value for the number of nearest neighbours $k$ considered per data point in deriving this maximal f-measure is $\mu=11.4$ with standard deviation $\sigma=9.8$ and median $m=7$. For normal points $\mu=11.6$, $\sigma=9.7$ and $m=12.5$, while for NN-outliers this is lower with $\mu=10.2$, $\sigma=10.4$ and $m=5$. Maximal $k$ having minimal density for both normal data and outliers was 24.

More general results computed across all available time-series for varying maximum $k$ are shown in Fig. 22 and Fig. 23.

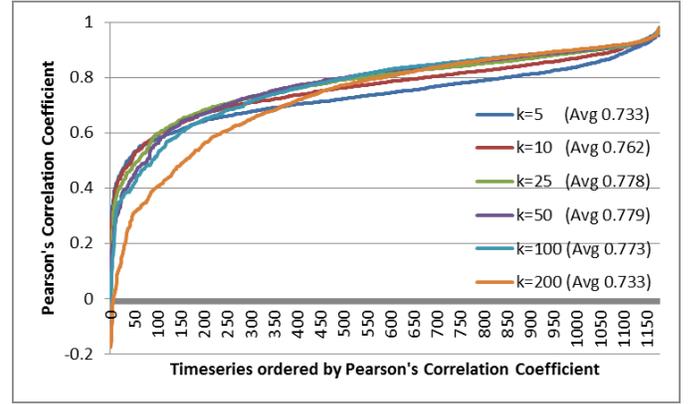

**Figure 23. Rank agreement to variable *k*-nearest neighbor**

**RQ6:** Is it possible to compute an exact finite specification for the *k*-nearest neighbour algorithm presented in RQ5?

Presume that this algorithm is trained on a two dimensional data set $\{p_i\}$, and assume that nearest neighbor considerations are with respect to this training set (as is the case for *iforest*). Further assume that some maximal inverse population density $D$ computed from this training set is deemed to be associated with all normal data points. This established radiuses $\{r_k : \pi r_k^2/k <= D\}$ these being the maximal distances that any normal data point may lie from $k$ or more training points.

Consider the *1*-nearest neighbour algorithm presented in RQ5. Since normal data points lie at most $r_1$ from one or more training points, the normal space is the union of all potentially overlapping circles $C_{1,i}$ of radius $r_1$ centered at $p_i$.

Generalizing to arbitrary $k$, the $C_{k,i}$ are formed by extending all $C_{1,i}$ to have radius $r_k$. The intersection of each $k$ distinct $C_{k,i}$ identifies one area of the data space where at least $k$ training points lie no further than $r_k$ from any point in this space. Each such intersection has a boundary formed by $\leq k$ individual arcs of the $C_{k,i}$ and so can be readily represented (at least to some given precision). Any data point lying within distance $r_k$ of $k$ or more training points must lie somewhere within or on the boundary of at least one such intersection. The union of all such intersections thus define the normal space for increasing $k \leq max$. The anomalous data space is the union of the compliment of each of these normal spaces, since a data point $d$ is anomalous if there exists a least one circle of radius $r_k$ centred at $d$ containing less than $k$ training points.

This construction can be used to create an algorithm that generates specifications that can then be directly referenced. Since specification offers significant benefits over detection, implementation of such an algorithm is certainly justifiable. It is interesting that an exact finite specification (predicated only on training data) can be obtained for two such different anomaly detection algorithms so easily, and it encourages research into generating specifications for other algorithms. This has historically been presumed to be very challenging, but we have demonstrated that that is not always so. Results for nearest-neighbour anomaly detection for up to 25 nearest neighbours are shown in Fig. 24.

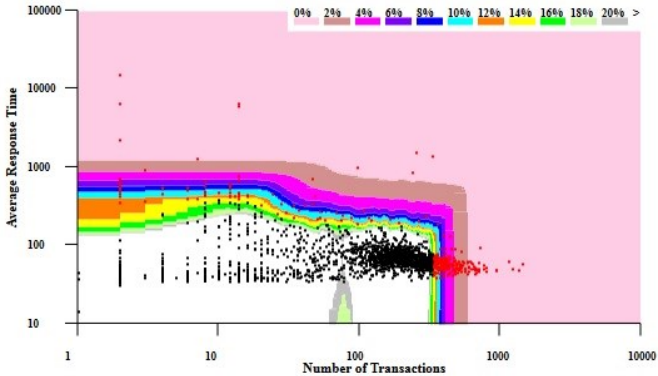

**Fig 24. Contour map for *k*-neighbor anomaly detection**

## 10. THREATS TO VALIDITY

All experiments were conducted on Visual Studio C++ (open source) code compiled in release mode. They were run on a 64 bit Windows 7 operating system using an Intel i7-2600 dual CPU, with each CPU having a clock speed of 3.4GHz. The machine had 8GB of RAM. Other activities were being concurrently undertaken. Execution times were obtained to the accuracy achievable using the Query Performance Counter interface.

Our *iforest* algorithm currently always identifies at least one anomaly in any non-empty set of data. It will do this even if there are no anomalous values in this data set. Post processing may therefore be required to remove from values specified as anomalies, those false positives that are subsequently deemed not to be anomalous.

The accuracy of this algorithm is largely dependent on how closely the estimated number of anomalies agrees with the number of anomalies in the data. Close agreement produces remarkably good results, while significant disagreement results in very poor precision and recall.

Reassuringly, repeatedly running this algorithm on unchanging data rapidly improves the guess as to percentage of data values that are likely to be anomalous. For example, the results in Table 1 make it clear that the expected number of anomalies in this repeatedly but randomly constructed data set is around 50, which is entirely consistent with approximately 1% of 5,000 values being by uniformly random construction anomalous.

It has not been demonstrated that the proposed algorithm is superior to all other anomaly detection algorithms (Luca 2014; Wu 2014). Any such comparison would be somewhat subjective. Consider RQ1 where three different anomaly detection algorithms are compared. We observe some overlap between the values detected as anomalous by the three algorithms, but we cannot (without ground truth about what constitutes an anomaly) say which anomaly detection algorithm is consequently best.

Experiment 11 provides perhaps as good a definition for an anomaly as any, in suggesting that an anomaly is a rare value that multiple independent evaluators would all agree was anomalous. And that experiment showed that the proposed algorithm is remarkably effective at discovering this type of anomaly. Experiment 12 where (in an unsupervised manner) the expected properties of CPU load are recovered, reinforces this conclusion, since the results shown in Figure 17 are quite consistent with actual knowledge about appropriate CPU loads.

RQ5 and RQ6 make it clear that the behaviour of the *iforest* algorithm is similar to *k*-nearest neighbour detection, at least on one large data set available to us.

The *iforest* algorithm performs poorly when anomalous regions about normal regions. In this case the algorithm tends to misspecify anomalous values as normal, and vica versa. This is a natural consequence of being an unsupervised algorithm that is given no instruction regarding the nature of anomalies. The effect is more pronounced in higher dimensions.

This algorithm is inherently stochastic, and leverages the law of large numbers. Results presented in this paper may not be statistically significant. Search path lengths will vary, since their length is determined by the randomly selected values used to train the algorithm, and the random partitioning performed. However, it is likely that such variations will only be significant where the definition of anomalous is already somewhat arbitrary.

The proposed extension to precisely determine all anomalous spaces (rather than to simply detect anomalies within these spaces) will identify the same anomalies as the original *iforest* algorithm, since this extension models within its decision tree precisely the decisions performed by the original algorithm in identifying anomalies, presuming that code is indeed implemented correctly. However it may be necessary to approximate multivariate results involving many dimensions to achieve satisfactory space and execution times.

The training tree and forest sizes are the only parameters this algorithm employs to discover anomalies. This paper uses sizes consistent with sizes recommended in an earlier paper. However we have not validated that these sizes are optimal.

No matter how accurately this algorithm isolates anomalies, the probability that it will not always do so increases as the number of data values being tested increases. False negatives and false positives will occur, and this should be anticipated in any application expoiting the described algorithm.

## 11. CONCLUSIONS AND FURTHER WORK

The isolation forest anomaly detection algorithm identifies anomalous data values by considering how rare they are and their distance from other more normal values. Its behavior is similar to an algorithm that identifies outliers having minimal population density with respect to some *k* nearest neighbors.

An extension to this unsupervised algorithm has been presented, which transforms it from one that can only identify values as normal or anomalous, into one that instead autonomously specifies all *regions* of a multi-variate data space as being either normal or anomalous. This permits tabulation and specification of what **would** be anomalous if observed, as well as detection of actually observed anomalies, with many resulting benefits. These have been described. Similar extensions can be applied to any algorithm that uses a metric associated with ordered trees or k-nearest-neighbor to identify anomalies or to otherwise classify data (Aryal 2014).

The enhanced *iforest* algorithm is simple, powerful, and effective. It can identify all anomalous data regions for one dimensional data in constant time, since in need only examine the training set to identify anomalous data spaces. It can then detect if any value is normal or anomalous, by examining if the region it is contained within is considered to be anomalous. This lookup operation can also be done efficiently in some small constant time. Since a value must be examined to be classified no algorithm could do better.

An *iforest* algorithm for specifying which multivariate hyper-rectangles in two or more dimensions are anomalous has also been presented. This algorithm includes precisely the earlier computed anomalous values within the anomalous data space that it identifies, given time and space, but can execute much faster if approximate results are acceptable. It remains to be seen how well this generalizable algorithm scales to more than three dimensions. It has been applied to a 3-dimensional color imaging problem with considerable success.

Access to the specification permits post-tailoring of rules regarding where anomalies arise, and has the potential to dramatically improve performance, both in time and space, by computing just once how the *iforest* algorithm will (if executed) behave, and thus what data regions it will identify as anomalous. These computed tabulated results may be exploited, as an alternative to expensive internal computation. This is of particular benefit in real time systems where strict usage constraints exist for both time and space.

Knowing the regions of a data space in which anomalies may occur permits an evolving profile to be constructed for distinguishing normal values from anomalous ones. This profile can be generated automatically, which is of benefit to engineers unfamiliar with the data being examined. The profile indicates the frequency with which anomalies occur (and where they occur) in real-time or windowed data, and so has application in the testing, monitoring and validating computer systems. The information about what constitutes an anomaly can be periodically recomputed, compared and/or manually adjusted as necessary, with resulting improvement in the overall anomaly detection process.

Since the anomaly detection process is entirely unsupervised it has applications in unsupervised autonomous specification and testing of both hardware and software, and may be incorporated into other autonomous algorithms.

The proposed algorithm has been tested on simulated one-, two- and three-dimensional multivariate data, and used to discover potential data anomalies in twelve years of data obtained hourly from a very large cloud environment, and nine months of data obtained hourly from a second different cloud provider. This algorithm has also been demonstrated to be useful in very diverse applications.

More work is needed to provide quantative measures of the reliability of the *iforest* algorithm for a given tree and forest size, and for different types of anomaly. Better techniques for predicting the number of anomalies in data and/or ways of isolating anomalous data would also be valuable.

The search path depths for normal and anomalous values behave as if they are derived from a mixture of two quite different distributions. If these two distributions could be separately identified, possibly by using the expectation maximization (EM) algorithm (Bilmes 1978), the optimal prediction for the number of anomalies would then be where these two probability density functions intersect, since values at this point would be equally likely to be derived from either distribution. This might be a fruitful avenue for further investigation.